\def\year{2021}\relax
\documentclass[letterpaper]{article} 
\usepackage{aaai21}  
\usepackage{times}  
\usepackage{helvet} 
\usepackage{courier}  
\usepackage[hyphens]{url}  
\usepackage{graphicx} 
\usepackage{natbib}  
\usepackage{caption} 
\title{An End-to-End Solution for Named Entity Recognition in eCommerce Search}
\author{
    Xiang Cheng,
    Mitchell Bowden,
    Bhushan Ramesh Bhange, \\
    Priyanka Goyal,
    Thomas Packer,
    Faizan Javed
    \\
}
\affiliations{

    The Home Depot \\ 
    2455 Paces Ferry Rd NW, Atlanta, GA 30339 \\

    \{xiang\_cheng, mitchell\_s\_bowden, bhushan\_ramesh\_bhange, faizan\_javed\}@homedepot.com

}

\begin{document}

\maketitle

\begin{abstract}
Named entity recognition (NER) is a critical step in modern search query understanding.  In the domain of eCommerce, identifying the key entities, such as brand and product type, can help a search engine retrieve relevant products and therefore offer an engaging shopping experience. Recent research shows promising results on shared benchmark NER tasks using deep learning methods, but there are still unique challenges in the industry regarding domain knowledge, training data, and model production. This paper demonstrates an end-to-end solution to address these challenges.  The core of our solution is a novel model training framework "TripleLearn" which iteratively learns from three separate training datasets, instead of one training set as is traditionally done. Using this approach, the best model lifts the F1 score from 69.5 to 93.3 on the holdout test data.  In our offline experiments, TripleLearn improved the model performance compared to traditional training approaches which use a single set of training data. Moreover, in the online A/B test, we see significant improvements in user engagement and revenue conversion.  The model has been live on homedepot.com for more than 9 months, boosting search conversions and revenue. 
Beyond our application, this TripleLearn framework, as well as the end-to-end process, is model-independent and problem-independent, so it can be generalized to more industrial applications, especially to the eCommerce industry which has similar data foundations and problems.
\end{abstract}

\section{Introduction}
The search engine at homedepot.com processes billions of search queries, serves tens of millions of customers, and generates tens of billions of dollars in revenue every year for The Home Depot (THD). One of the most fundamental challenges in our search engine is to understand a search query and extract entities, which is critical to retrieve the most relevant products.  This task can be framed as named entity recognition (NER) which is a common information retrieval task to locate, segment, and categorize a pre-defined set of entities from unstructured text.
Since its introduction in the early 1990s, NER has been studied extensively and is evolving rapidly \cite{sekine09nyu, yadav2019survey}, especially after the adoption of deep learning and related techniques in recent years \cite{yadav2019survey}.

However, there is a gap between academic research and industrial applications of NER.  Recent research works often use the latest neural architectures \cite{yadav2019survey} and language models \cite{glove2014,bert2018,elmo2018} to improve performance on popular NER shared tasks and datasets \cite{conll2003}. The focused outcome is often marginal improvement of F1 scores, while the feasibility in real-world applications is not often considered.

\subsection{Recent Research in NER}
In recent years, deep-learning-based NER together with embeddings has become increasingly popular in the research community. Collobert et al. proposed the first neural network architecture for NER \cite{collobert2008, li2020survey} and later experimented pre-trained word embeddings \cite{collobert2011cnn} as model features. Since then various neural architectures and word representations have been studied \cite{yadav2019survey}. The most popular approach is to use recurrent neural networks (RNN) over word, sub-word, and/or character embeddings \cite{yadav2019survey,lee2017lstm}. Long Short Term Memory (LSTM) and its variants (e.g. Gated Recurrent Unit, i.e. GRU) are the most common neural architectures for NER as well as for sequence tagging \cite{huang2015lstm,ma2016cnn,lample16neural, lee2017lstm}. Recently, transformer-based language models \cite{vaswani2017attention, bert2018} have been tested on benchmark NER tasks and claim the state-of-art performance \cite{liu2019bertner, li2020survey}.

It is exciting to see the booming research progress based on modern deep learning methods and latest language models. However, industrial applications still seem being left behind due to unique challenges as described below. 

\begin{table*}[tbp]
    \centering
    \begin{tabular}{c|c}
        \hline
    {\small Legacy NER Entities} & {\small True Entities}  \\
    \hline
    \hline
    $\overbrace{\texttt{fridge no}}^{O} \ \overbrace{\texttt{ice maker}}^{PRD}$ 
    & $\overbrace{\texttt{fridge}}^{PRD} \ \overbrace{\texttt{no ice maker}}^{O}$ \\
    \hline   
    
    $\overbrace{\texttt{weed eater}}^{BRD} \ \overbrace{\texttt{light}}^{PRD} \ \overbrace{\texttt{weight}}^{O}$
    &  $\overbrace{\texttt{weed eater}}^{PRD} \ \overbrace{\texttt{light weight}}^{O}$ \\
    \hline
    
    $\overbrace{\texttt{cosco}}^{BRD} \ \overbrace{\texttt{table}}^{PRD} \ \overbrace{\texttt{and chair set}}^{O}$
    &  $\overbrace{\texttt{cosco}}^{BRD} \ \overbrace{\texttt{table and chair set}}^{PRD}$ \\
    \hline
    \end{tabular}
    \caption{The examples where legacy NER mislabeled entities. The first example has two product types (i.e. \texttt{"ice maker"} and \texttt{"fridge"}). \texttt{"weed eater"} is ambiguous because it could be either a brand or a product type.  For the third one, \texttt{"table and chair set"} is not in the existing product taxonomy.}
    \label{table:nerproblem}
\end{table*}

\subsection{Industry Applications and Challenges}
Compared to the large amount of research on deep-learning-based NER, there are relatively few publications that explore its applications in industrial settings, and fewer still that address the task of performing NER as a step in eCommerce search query understanding.

Guo et al. \cite{guo2009nermicrosoft} apply a probabilistic topic model, Weakly Supervised Latent Dirichlet Allocation, to identify four types of entities from commercial web search queries containing single named entities.  It is not clear if this approach was fully evaluated in their online setting.  

Putthividhya and Hu \cite{ner2011ebay} use statistical sequence models to recognize entities (product brands and designers).  They use character $n$-gram similarity scores to resolve entities to canonical forms.  The evaluation appears to be offline only, using around 2K clothing product listings.   

Cowan et al. \cite{cowan15expedia} use a conditional random field (CRF) to recognize three types of entities in travel search queries.  3.5K manually labeled queries are used in the evaluation.  The NER system was used in production, though the evaluation appears to be offline only.  

More \cite{more16walmart} uses CRFs and a voted-perceptron-based Markov model--both forms of supervised learning.  They utilize a large set of unlabeled data by using regular expressions to label training data for distant supervision.  They also evaluate the value of brand name recognition in production.  

Wu et al. \cite{wu2017google} identify product attributes intended by user queries 
by treating this as a multi-label text categorization problem.  They train character- and word-level bidirectional-LSTMs (BiLSTM) jointly with a product-attribute auto-encoder, using implicit user feedback and no hand-labeled training data.  

Majumder et al. \cite{majumder2018walmart} evaluate multiple deep recurrent networks by extracting brand names from product titles.  They mention other named entity types they may have also evaluated in production but it is not clear whether these results were cherry-picked from a larger experiment.  

Wen et al. \cite{wen2019ebay} describe a production NER system in eCommerce for extracting entities from search queries using a process that improves the accuracy of extracted information over its distant-supervision training data which comes from a legacy NER system based on logistic regression.  They do not specify which entities they evaluated nor what their performance was in absolute terms.  They mention that their approach is efficient in terms of human labeling cost.  However, it is dependent on distant supervision from a legacy NER system that ``is built upon a lot of domain expertise-aided feature engineering".  Therefore, it is impossible to determine how much labeling or engineering cost the current system relies on as their description is not truly end-to-end.  

Among these papers, three papers \cite{wu2017google, majumder2018walmart, wen2019ebay} explore deep learning for NER in eCommerce search.  Two papers \cite{wu2017google, wen2019ebay} perform an evaluation on queries in production.  Only four papers \cite{ner2011ebay, more16walmart, wu2017google, wen2019ebay} leverage a large number of unlabeled queries available to commercial search engines.

Despite the above progress, we believe the following challenges remain. 
\begin{enumerate}
    \item Industrial applications require custom, domain-specific knowledge and training data, covering the full extent of entities of different types, including representative examples of noisy queries (spelling errors, abbreviations, etc.) and noisy intent signals from conversion (purchase) events.  
    \item Machine learning models, especially deep learning models, usually require a large amount of high quality training data, which is often time-consuming and sometimes impossible to acquire. 
    
    \item Productionization is challenging as deep learning models are computationally expensive to train and execute in a real-time application such as eCommerce search.  
\end{enumerate}

To address these challenges, this paper demonstrates an end-to-end solution (see Sec. \ref{sec:secframework}), including preparing training data in a scalable manner, iteratively training the model using the TripleLearn framework, and deploying the model in production for real-time usage in our eCommerce search engine. 


\subsection{NER at The Home Depot} \label{sec:thdner}
THD is a leading home improvement retailer, and this domain is rich in entity types and relationships. For example, THD has more than 14,000 brands, 11,000 product types, and 3 million items. In our search engine, NER is a vital but challenging step to extract the key entities. 

The legacy NER system in production used pre-defined taxonomies of brands and product types, recognized using a sequential greedy exact match. Beginning with brands and then product types, the longest matching token sequence was labeled as the corresponding entity. However, this approach was far from satisfactory, and common challenges included queries containing multiple product types but only one shopping product type, ambiguity between product types and brands, and new product types not in pre-defined taxonomy. Table \ref{table:nerproblem} has examples to demonstrate these challenges. These mislabels were often closely related to the context in the query and specific entity values (brands or product types), so it was hard to fix at scale in the legacy NER system. Deep learning models could be good candidates to handle this kind of complexities \cite{wu2017google, majumder2018walmart, wen2019ebay}, which is the type of model we will use. 

\hfill

The rest of this paper is organized as follows.  Sec. \ref{sec:secframework} defines the problem and describes the whole process including the TripleLearn framework in detail.  Sec. \ref{sec:expresults} covers offline experiments and results while Sec. \ref{sec:prod} covers online deployment and measured impact to business. In Sec. \ref{sec:dis}, we discuss the TripleLearn training framework and future work.

\section{Technical Design Overview} \label{sec:secframework}

In this section, we first define the problem, then elaborate the end-to-end process including the TripleLearn framework, and describe the selected model architecture in the last subsection.

\subsection{Problem Definition}

Given a user search query as a sequence of word tokens, the primary task of NER is to identify the important entities.  This task is formulated as a sequence tagging problem using beginning-inside-outside (BIO) tagging format.  In this paper, we focus on the two most important entities for eCommerce: brand (\texttt{BRD}) and product type (\texttt{PRD}).  Therefore, two entities are translated into five labels as shown in Table \ref{table:label}.  The general process to label the entities in a search query is shown in Table \ref{table:def}.  The key step is to build a machine learning model for sequence tagging in Step 3.  The model evaluation metric is the exact-match F1 score \cite{conll2003} where only the correct prediction of the whole entity is considered as a true positive.  The baseline is the legacy NER system in production as described in Sec. \ref{sec:thdner}.

\begin{table}[t]
    \centering
    \begin{tabular}{@{}l|l@{}} 
    \hline
    Label & Description \\
    \hline
    \hline
    \texttt{B-BRD} & beginning of a brand \\ \hline
    \texttt{I-BRD} & inside of a brand \\ \hline
    \texttt{B-PRD} & beginning of a product type \\ \hline
    \texttt{I-PRD} & inside of a product type \\\hline
    \texttt{O} & outside \\\hline
    \end{tabular}
    \caption{The five labels in this NER problem.}
    \label{table:label}
\end{table}

\begin{table}[t]
    \small
    \centering
    \begin{tabular}{@{}l|l@{}}
    \hline
    Step & Example \\
    \hline
    \hline
    1. search query & \texttt{LG washer mini}\\ \hline
    2. query preprocessing & \texttt{["lg", "washer", "mini"]} \\ \hline
    3. NER & \texttt{["B-BRD", "B-PRD", "O"]}\\ \hline
    4. output & \texttt{\{}\\
        &\quad  \texttt{"brand": "lg",} \\
        &\quad \texttt{"product": "washer"}\\
        & \texttt{\}} \\\hline
    \end{tabular}
        \caption{General steps to label entities in a search query.}
    \label{table:def}
    \normalsize
\end{table}

\subsection{End-to-End Process with TripleLearn} \label{sec:framework}

\begin{figure*}[t]
    \centering
    \includegraphics[width=1.82\columnwidth]{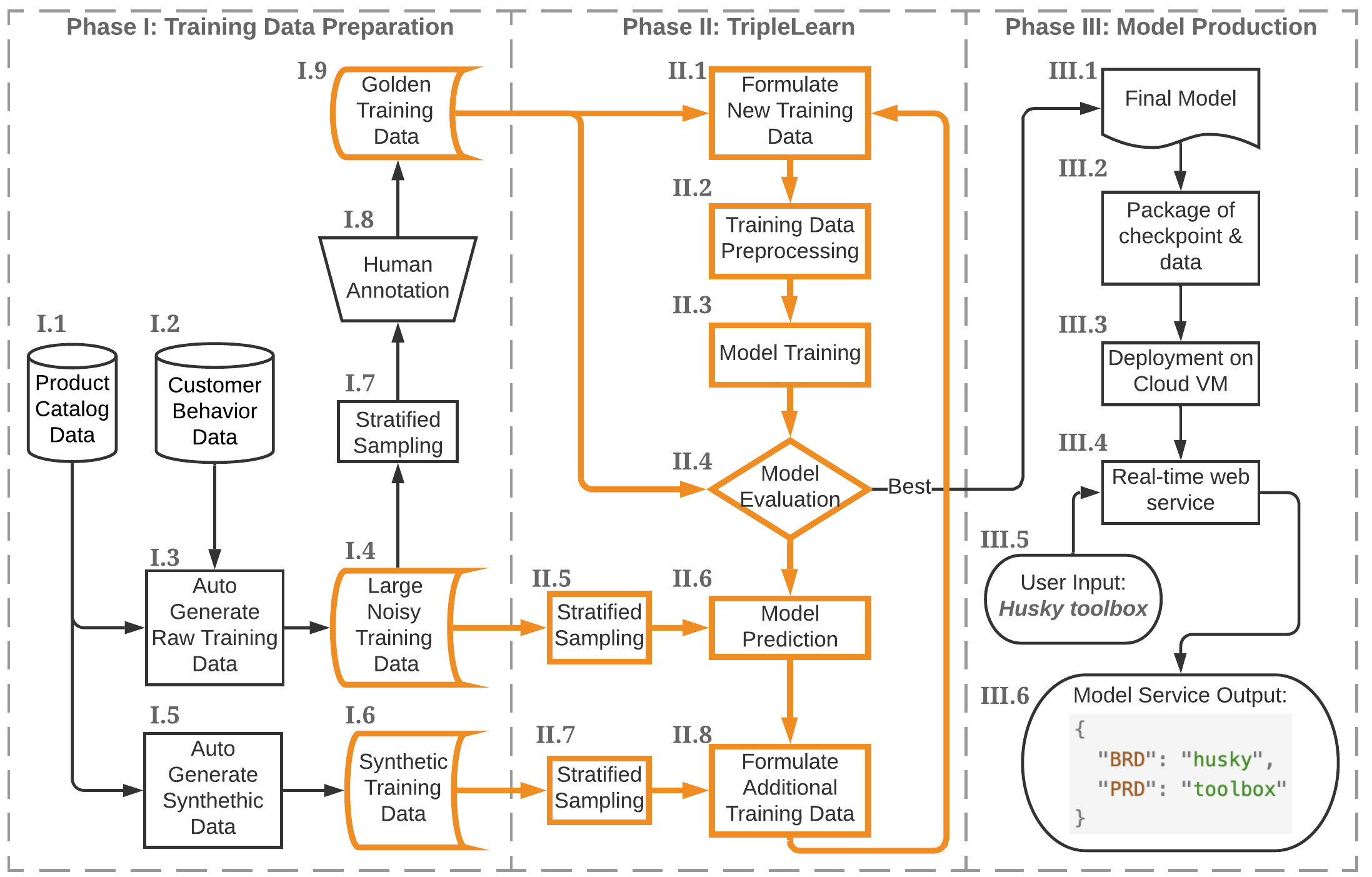} 
    \caption{The end-to-end solution with TripleLearn framework highlighted in bold and orange. Phase I prepares three sets of training data for TripleLearn, and Phase II is the iterative training process of TripleLearn. The last phase is model production.}
    \label{fig:flowchart}
\end{figure*}

Our end-to-end process from training data creation to model deployment is shown in Fig. \ref{fig:flowchart}.  The core is the TripleLearn framework in Phase II.  Each phase is described as follows. 

\subsubsection*{\textbf{Phase I: Training Data Preparation.}}

An ideal set of training data for deep learning models should have three characteristics: 

\begin{enumerate}
  \item large volume to train a large number of model parameters;
  \item high quality labels to provide correct supervision; 
  \item high coverage of label values (i.e. all brands and product types here) to potentially recognize all of them. 
\end{enumerate}

However, it is often too time-consuming to prepare one set of such training data.  We find that it is more realistic to prepare three separate datasets to meet these requirements.  These three sets of training data are prepared as described below. 

The starting point are two foundational datasets in an eCommerce business: the product catalog (node \texttt{I.1} in Fig. \ref{fig:flowchart}) and customer behavior data (node \texttt{I.2}).  Product catalog data is the ground truth for the key entities of all products sold on homedepot.com which have more than 14K unique brands and 11K unique product types.  Customer behavior data stores customers' shopping journeys, including searches, impressions, clicks, and purchases.  

Firstly, a large amount of training data (\texttt{I.4}) is automatically generated using rule-based algorithms (\texttt{I.3}) by matching the tokens in product catalog (\texttt{I.1}) and customer behavior data (\texttt{I.2}).  This dataset is large (2.7M queries) but noisy due to noisy customer behavior data and imperfect matching algorithms, but it can still capture the variety of patterns in real customers' search queries.

Secondly, to prepare a set of high quality ``golden'' data (\texttt{I.9}), we sample a small amount (16K) from \texttt{I.4} for manual annotation.  To avoid bias or over-fitting our model on a small number of patterns, we stratified sampled the data by entity sequence patterns.  Here, a pattern is defined as a 
series of consecutive entities.  
See Table \ref{table:pattern} for the top four patterns and examples.  The reason is that we find that our model is sensitive to the order of entities in the training data, which makes sense for a sequence tagging model.  This set of golden data is vital to start the model training in TripleLearn (Phase II) and to measure the real model performance.  

Thirdly, a rule-based algorithm (\texttt{I.5}) creates a set of synthetic data (\texttt{I.6}) using all the brands and products in our product catalog (\texttt{I.1}).  The synthetic pattern is simple, such as brand-only queries (e.g. \texttt{"samsung"} as a query) and product-type-only queries (e.g. \texttt{"washer"} as a query).  All the distinct brands and products are included so that the model can potentially recognize them all.

The outputs of Phase I are three datasets to meet these three requirements of an ideal training dataset.  Table \ref{table:datasets} shows corpus statistics and entity distributions.  These three datasets are the training data for TripleLearn as described next.

\begin{table}[t]
    \centering
    \begin{tabular}{l|l}
        \hline
    {\small Top Sequential Pattern} & {\small Example}  \\
    \hline
    \hline
    \texttt{BRD+O+PRD} & $\overbrace{\texttt{milwaukee}}^{BRD} \ \overbrace{\texttt{cheap}}^{O} \ \overbrace{\texttt{drill}}^{PRD}$ \\
    \hline
    \texttt{BRD+O+PRD+O} & $\overbrace{\texttt{ge}}^{BRD} \ \overbrace{\texttt{7.4 cu ft}}^{O} \ \overbrace{\texttt{dryer}}^{PRD} \ \overbrace{\texttt{gas}}^{O}$ \\
    \hline
    \texttt{BRD+PRD+O} & $\overbrace{\texttt{behr}}^{BRD} \ \overbrace{\texttt{paint}}^{PRD} \ \overbrace{\texttt{discount}}^{O}$ \\
    \hline
    \texttt{O+PRD+O} & $\overbrace{\texttt{bronze}}^{O} \ \overbrace{\texttt{faucet}}^{PRD} \ \overbrace{\texttt{pull down}}^{O}$ \\ 
    \hline
    \end{tabular}
    \caption{The examples of the top four sequential patterns in the large volume of training data (\texttt{I.4}). }
    \label{table:pattern}
\end{table}

\begin{table}[t]
    \centering
    \begin{tabular}{@{}l|r|r|r@{}}
    \hline
    {\small } & { Large (I.4)} & {Synthetic (I.6)} & { Golden (I.9)} \\
    \hline
    \hline
    Query & 2,737,399 & 23,710 & 16,915  \\ \hline
    Token & 16,914,607 & 48,568 & 89,692 \\ \hline
    \texttt{BRD} & 2,509,132 & 14,058 & 14,915 \\ \hline
    \texttt{PRD} & 2,694,413 & 9,652 & 14,774 \\ \hline
    \end{tabular}
    \caption{Corpus statistics.  The numbers of queries, tokens, and entities in the three training datasets for TripleLearn.
    }
    \label{table:datasets}
\end{table}

\subsubsection*{\textbf{Phase II: TripleLearn }}

TripleLearn is the core of this end-to-end solution.  In this phase, we iteratively train the model, cumulatively add more training data (i.e. more brands, more product types, and more patterns), and incrementally improve the model performance.

The first iteration starts with the golden data (\texttt{I.9}).  15\% of the golden data is randomly held out as test data; the rest is randomly split into training (90\%) and validation (``dev") data (10\%).  Before the training, the training data is pre-processed to balance the labels (\texttt{II.2}), which requires domain knowledge.  For example, we identify 50 tokens which can be either a brand or a product type, such as \texttt{"instant pot"}, \texttt{"cutter"}, and \texttt{"anchor"}.  Such queries are balanced by oversampling entities with fewer queries so that no bias is generated due to skewed data distribution.

Then the model is trained (\texttt{II.3}) until the F1 score on the validation set stops improving. The next step is to evaluate the model (\texttt{II.4}) on the test data to see whether this iteration has a better performed model. If the current iteration's model is worse than the previous one, we stop the iterative training and select the best model from all the previous iterations. If it is better, we continue the iterations by stratified sampling (\texttt{II.5}) from \texttt{I.4} as done in \texttt{I.7}, for model inference (\texttt{II.6}).  If the prediction on a query matches the noisy labels in \texttt{I.4}, we pass the query to the next step as additional training data.  The idea is to reduce the noise in training data by looking for consensus between the noisy labels produced by \texttt{I.3} and the noisy labels predicted by \texttt{II.6}.  

Similarly, we sample (\texttt{II.7}) synthetic data (\texttt{I.6}) to increase the coverage of brands and product types.  The goal is to cover all possible entity values after a few iterations.  In the end, we formulate new additional training data from \texttt{II.6} and \texttt{II.8} for the next iteration.

A question we had during early development was whether this iterative self-training process would accumulate errors, thereby biasing itself toward erroneous patterns of predictions by \texttt{II.6}.  We reduce the chance of the model drifting by stratified sampling by label sequence patterns in steps \texttt{I.7}, \texttt{II.5}, and \texttt{II.7}.  Beyond that, the final judge of this iterative process is the F1 score on the held-out test set.  In our experiments (Sec. \ref{sec:expresults}), we see that this iterative process can indeed improve the model performance iteration by iteration.  More results are shown in Sec. \ref{sec:expresults}.

\subsubsection*{\textbf{Phase III: Model Production.}}
Model production is an indispensable step to use the model in THD search engine. This phase starts with the best model selected from TripleLearn (Phase II). In node \texttt{III.2}, the neural graph, weights, and required data (vocabulary, word embeddings, etc) are packed for the next step. Then the model package is deployed on a cloud virtual machine (\texttt{III.3}). To use it in a real-time search engine, in node \texttt{III.4}, the other necessary components are also added to be able to parse a raw user search queries and to convert NER model prediction into machine readable outputs, which enables the deep learning model as a real-time web service. The sample input and output of this model service are shown in \texttt{III.5} and \texttt{III.6} in Fig. \ref{fig:flowchart}. More details on deployment and testing are described in Sec. \ref{sec:prod} (Productization).




\subsection{Selected Model Architecture}

\begin{figure}[t]
    \centering
    \includegraphics[width=\columnwidth]{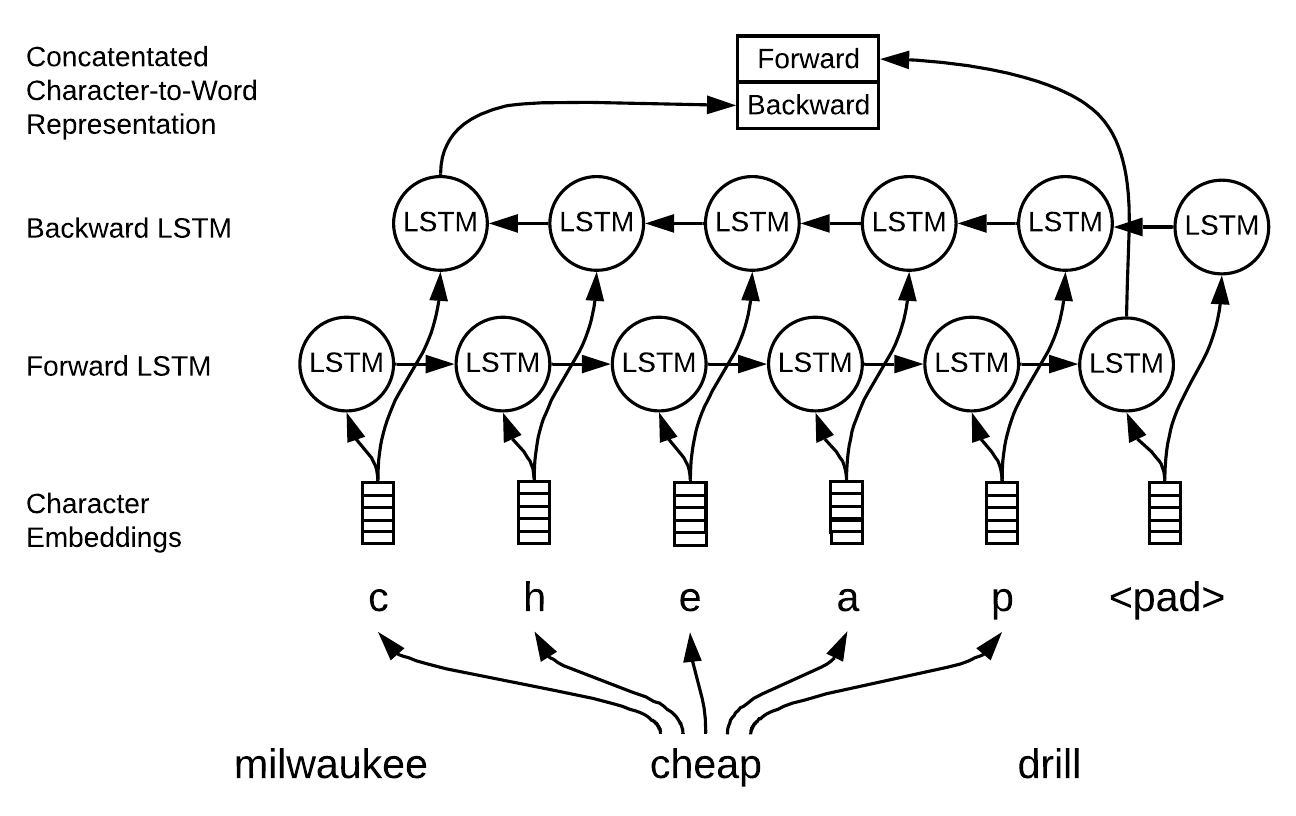}
    \caption{The character-to-word subgraph in our model.  Both forward and backward word representations are learned from character embeddings using a BiLSTM layer.}
    \label{fig:lstm}
\end{figure}

\begin{figure}[t]
    \centering
    \includegraphics[width=\columnwidth]{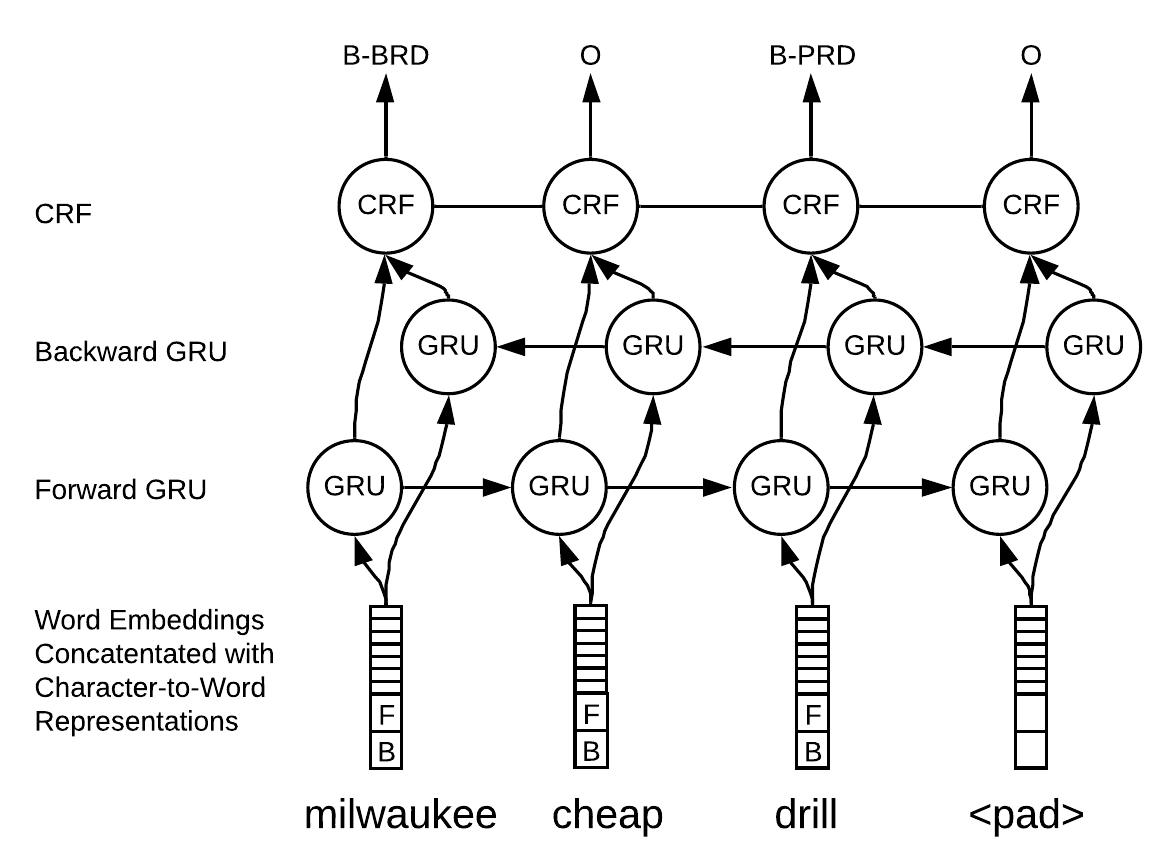} 
    \caption{The word-to-label-sequence subgraph.  Forward and backward character-based embeddings are concatenated to word embeddings as input to the bidirectional GRU (BiGRU) which in turn provides features for the CRF.}
    \label{fig:gru}
\end{figure}

After experimenting multiple neural architectures (Sec. \ref{sec:expresults}), our selected model is based on the bidirectional RNN-CRF, a popular approach to sequence tagging \cite{yadav2019survey, lee2017lstm}.  Recent works helped us to finalize the architecture.  Huang et al. first demonstrated that BiLSTM-CRF can effectively use both left and right context as well as the statistical sequential dependency among token labels \cite{huang2015lstm}.  Lample et al. \cite{lample16neural} further showed that BiLSTM-CRF with pre-trained word embeddings and character embeddings performed the best on CoNLL-2003 \cite{conll2003}.  The GRU, a simplified variant of the LSTM, has also shown state-of-the-art performance \cite{lee2017lstm}. After numerous experiments (see details in Sec. \ref{sec:expmodel}), our final neural architecture is a BiGRU-CRF with a BiLSTM subgraph for character-level embedding, as shown in Figures \ref{fig:lstm} and \ref{fig:gru}. This neural architecture is implemented using Tensorflow.

\section{Model Experiments} \label{sec:expresults}

We systematically run experiments on the TripleLearn framework, model architectures, and embeddings.  The evaluation metric is the micro-averaged F1 score which is commonly used in NER tasks. Because the number of the two entities are roughly balanced, so the macro-averaged F1 is also consistent with the micro-F1. The evaluation dataset is the holdout test data which is the 15\% random sample of the golden data in Phase I of Fig \ref{fig:flowchart}.

\subsection{TripleLearn Framework} \label{sec:exptriple}

\begin{figure}[t]
    \centering
    \includegraphics[width=0.72\columnwidth]{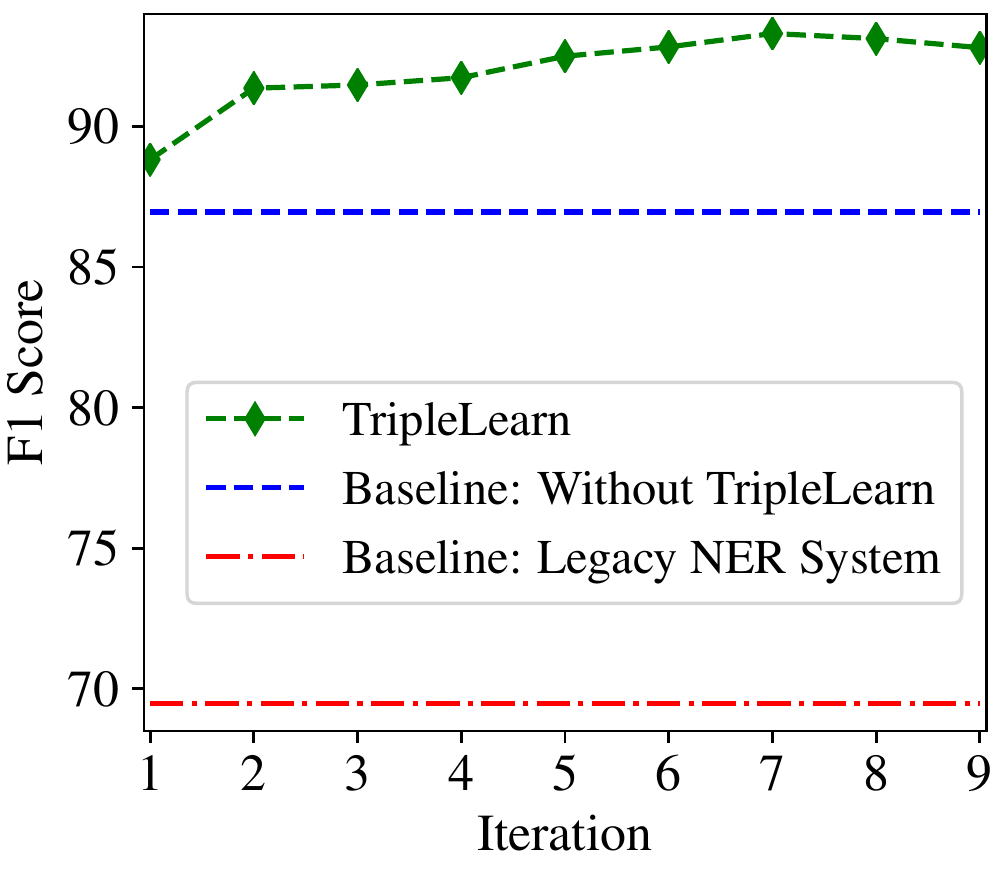}
    \caption{F1 score on the test data of TripleLearn at each iteration (green curve on the top) and two baselines.  The red line on the bottom is for the legacy NER system.  The blue line in the middle is for one-pass training using all the data (I.4 + I.6 + I.9 in Fig. \ref{fig:flowchart}).}
    \label{fig:iterperf}
\end{figure}

The experimental results show that TripleLearn (highlighted in bold in Fig. \ref{fig:flowchart}) iteratively improves the model performance on the test data as shown in Fig. \ref{fig:iterperf} and Table \ref{table:iter}. As we iteratively add more training data, the coverage of brands and product types increases and reaches 100\% at iteration 7, which is important for this model to be able to potentially recognize all brands and product types. The F1 scores also saturate after iteration 7. 

The advantage of TripleLearn is justified in three aspects. Firstly, the F1 score improves from 87.1 at iteration 1 to 93.3 at iteration 7, which demonstrates that it can iteratively improve the model.  Secondly, compared to simply training the model once using all the data (Fig. \ref{fig:iterperf} blue horizontal line), this iterative process performs better in every iteration.  Lastly, compared to the most important baseline, the legacy NER system in production, our best model significantly increases the F1 score (from 69.5 to 93.3), which shows the superiority of the model and justifies an A/B test.  The model at iteration 7 has the best F1 score and also covers all the brands and product types, so it is selected for production.

\begin{table}[t]
    \centering
    \begin{tabular}{@{}c|rrrrr@{}}
        \hline
        iter. & training &  {\small unq. BRD} & {\small unq. PRD} & {\small dev F1} & {\small test F1} \\
        \hline
        \hline
        1 & 14,378 & 3400 & 3239 & 89.32 & 88.82 \\
        2 & 19,911 & 3936 & 3768 & 95.81 & 91.36 \\
        3 & 31,510 & 4402 & 4216 & 96.52 & 91.47 \\
        4 & 57,379 & 5374 & 5131 & 97.70 & 91.73 \\
        5 & {\small 115,564} & 7544 & 7004 & 98.68 & 92.49 \\
        6 & {\small 241,629} & 12,082 & 10,305 & 99.19 & 92.83 \\
        \textbf{7} & {\small 484,390} & 14,102 & 11,111 & \textbf{99.49} & \textbf{93.30} \\
        8 & {\small 992,823} & 14,102 & 11,111 & 99.66 & 93.12 \\
        9 & {\small 2,089,573} & 14,102 & 11,111 & 99.81 & 92.79 \\
        \hline
    \end{tabular}
    \caption{The F1 scores, the numbers of training queries, unique brands, and unique product types by iteration.  
    }
    \label{table:iter}
\end{table}

A disadvantage of this process is that it takes a longer time because we have to train multiple iterations. On average, the whole process with nine iterations takes about 20 hours on a GPU machine (NVIDIA Tesla K80) using our training data. However, the training is an offline process, and we value the quality of the model more than the offline training cost.

Considering the long training time, most of the experiments below are tested on golden data only (iteration 1), which is much faster. However, from the finished iterative model trainings, we find that the model performance at iteration 1 is already a good indicator of the final model performance, which can justify the experimental findings using the golden data in the next subsections.

\subsection{Tested Models}  \label{sec:expmodel}

\begin{table}[t]
    \centering
    \begin{tabular}{@{}c|c|cc@{}}
        \hline
        models & char emb. & dev F1 & test F1 \\
        \hline
        \hline
        BiLSTM & 
            \begin{tabular}{c} No \\ Yes \end{tabular} & 
            \begin{tabular}{c} 85.77 \\ 86.99 \end{tabular} & 
            \begin{tabular}{c} 85.05 \\ 86.23 \end{tabular} \\
        \hline
        BiLSTM-CRF & 
            \begin{tabular}{c} No \\ Yes \end{tabular} & 
            \begin{tabular}{c} 87.69 \\ 88.57 \end{tabular} & 
            \begin{tabular}{c} 86.72 \\ 88.44 \end{tabular} \\
        \hline
        BiGRU & 
            \begin{tabular}{c} No \\ Yes \end{tabular} & 
            \begin{tabular}{c} 85.42 \\ 86.53 \end{tabular} & 
            \begin{tabular}{c} 85.57 \\ 87.09 \end{tabular} \\
        \hline
        \textbf{BiGRU-CRF} & 
            \begin{tabular}{c} No \\ \textbf{Yes} \end{tabular} & 
            \begin{tabular}{c} 87.12 \\ \textbf{88.71} \end{tabular} & 
            \begin{tabular}{c} 87.04 \\ \textbf{88.82} \end{tabular} \\
        \hline
        BERT & N/A & 83.22 & 82.51 \\
        \hline
        BERT-CRF & N/A & 83.93 & 83.10 \\
        \hline
    \end{tabular}
    \caption{Six neural architectures tested on golden data, with and without layers of character-based embedding if applicable.}
    \label{table:models}
\end{table}

We test six neural architectures as shown in Table \ref{table:models}. Each neural layer is tested and discussed below.


The character-based embedding is produced by a BiLSTM layer \cite{lample16neural} as shown in Fig. \ref{fig:lstm}.  It essentially extracts features for each word using a character-level model.  With character-based embedding, the F1 score is consistently better as shown in Table \ref{table:models} using golden data for training and testing.  We believe the reason is that it can help bridge the gap between common query words and uncommon query words that contain common sub-word character sequences. 

We also compare BiGRU and BiLSTM, with BiGRU showing comparable or even better performance.  We select BiGRU because it has 25\% fewer parameters and is thus faster to train and execute compared to BiLSTM.

In addition to the RNN-based neural models, we also tested transformer-based BERT (uncased large) \cite{bert2018}, but the performance is less satisfactory as shown in Table \ref{table:models}, either with a CRF layer or with a simple softmax layer. It may be because our search queries are domain-specific and have very different patterns comparing to the BERT training corpus (i.e. Wikipedia and books). Further improvements may require fine-tuning using domain specific corpus and/or additional neural layers (e.g. BERT-LSTM-CRF), which requires significantly more efforts in terms of both training and deployment.

The CRF layer is helpful to predict the most likely entity sequence and to forbid invalid sequence transitions, such as \texttt{B-BRD$\rightarrow$I-PRD} and \texttt{O$\rightarrow$I-BRD}. Our results in Table \ref{table:models} also show that the CRF layer consistently improves F1 score.

\subsection{Word Embeddings}

Word embedding is an important component and simplifies feature engineering \cite{yadav2019survey}.  We test pre-trained Word2vec embedding \cite{mikolov2013efficient}, pre-trained GloVe \cite{glove2014}, custom  Word2vec, and custom GloVe embedding.  The custom embeddings are trained using the top 50 million search queries and 3 million product titles on homedepot.com.  

\begin{table}[t]
    \centering
    \begin{tabular}{c|ccc}
        \hline
        embedding & \begin{tabular}{@{}c@{}}similar \\ words \end{tabular} & \begin{tabular}{@{}c@{}}vocab. \\ coverage \end{tabular}  & F1\\
        \hline
        \hline
            \begin{tabular}{@{}c@{}}pre-trained \\ GloVe \end{tabular}   &
            \begin{tabular}{@{}c@{}}chicago \\ detroit \\ minneapolis \end{tabular} & 
            39.8\% & 87.56\\ \hline
            \begin{tabular}{@{}c@{}}pre-trained \\ Word2vec \end{tabular}   & 
            \begin{tabular}{@{}c@{}} springfield \\ harvey \\ wisconsin \end{tabular} & 
            15.5\% & 83.99 \\ \hline
            \begin{tabular}{@{}c@{}}custom \\ GloVe \end{tabular}   &
            \begin{tabular}{@{}c@{}} m18 \\ dewalt \\ drill \end{tabular} & 
            99.9\% & 87.58 \\ \hline
            \begin{tabular}{@{}c@{}}\textbf{custom} \\ \textbf{Word2vec} \end{tabular}   & 
            \begin{tabular}{@{}c@{}} dewalt \\ makita \\ ridgid \end{tabular} & 
            \textbf{99.9\%} & \textbf{88.82} \\ \hline
    \end{tabular}
    \caption{Top three similar words for \texttt{"milwaukee"}, vocabulary coverage of all unique words in training data, and F1 score on the test data of the four word embeddings.  
    }
    \label{table:emb}
\end{table}

We select custom Word2vec embedding in our final model for three reasons.  Firstly, domain-specific word embedding is a better semantic representation in terms of similar words.  Table \ref{table:emb} shows the top three similar words for \texttt{"milwaukee"} which is a city but also a popular brand in the home improvement domain. Secondly, the vocabulary coverage is much higher for the custom-trained embeddings.  Lastly, as shown in Table \ref{table:emb}, the model performance is also better than the pre-trained embeddings.




\section{Productionization} \label{sec:prod}
Productionization is the key to delivering a real-world impact, and there are two correlated challenges, speed and cost. Firstly, the model execution has to be fast enough to serve thousands of queries per second and help retrieve search results within a time limit. The speed also has a direct impact on user satisfaction and conversion rate. Secondly, the cost of serving the model has to be reasonably low to justify the return on investment. More details are explained below.

\subsection{Deployment}

The first challenge is speed, and a practical solution is to leverage an existing platform with customized optimization. Specifically, the model is deployed using Tensorflow Serving, a flexible and high-performance model serving system by Google, on the Google Cloud Platform (GCP).  This can provide a stable environment but still not an optimized one. Thus, we customize the optimization by reducing the servable model size which has a direct impact on the model inference time. This involves converting the variables in a Tensorflow checkpoint into constants stored directly in the model graph, stripping out unreachable parts of the graph, folding constants, folding batch normalizations, removing training and debug operations, etc. In our experience, the optimized model has a significantly smaller size, faster loading, and faster inference.  

The other challenge is cost. Serving the deep learning model on a GPU machine would be fast but also much more expensive. We manage to optimize the model for CPU to meet our performance requirement and deploy it on a GCP virtual machine instances with custom CPU machine type (4 vCPUs, 3.75 GB memory). This deployment automatically scales to real-time traffic, leading to a very cost-effective solution.  

The model has been live in production for more than 9 months, and we see that the model service can serve 99\% of the search queries within 26 milliseconds.  This is highly satisfactory for our search engine. This concludes a successful engineering deployment of the model.

\subsection{Usage, A/B Test, and Business Impact}

The deep learning model service is used in our search engine serving a high volume of search queries in real time. The extracted entities (brand and product types) are used to retrieve items in the search inverted index as well as serving as an additional input for ranking.

In the A/B test, we tested the NER model service against the legacy NER system with equally random traffic split at homedepot.com.  With millions of search visits from real online shoppers, we saw a significant improvement in both click-through-rate and revenue conversion per search.  The estimated annual business impact is \$60M in incremental revenue. The cost is minimal comparing to its benefit, so this project is yielding a high return on investment. Detailed metrics measured from the A/B test are confidential and not disclosed here. 

After the A/B test, this model has replaced the legacy NER system and has been live on homedepot.com for more than 9 months, serving millions of real customers and boosting search conversions.

\subsection{Maintenance} \label{sec:prodmaintain}
There is no model that works well forever. It is especially true in the eCommerce scenario where new products with new brands and/or new product types are added regularly. Therefore, we also need to refresh the model regularly to reflect the changes.

The TripleLearn framework makes it easy to refresh the model because we only need to incrementally update a small fraction of the three datasets to retrain the model. For example, we recently refreshed the model to improve the performance on short queries that have more average traffic than long queries. In this case, we only added 5\% additional short-query-training-data to the golden dataset (i.e. I.9 in Fig. \ref{fig:flowchart}) to retrain the model. In both the offline evaluation and A/B test, the refreshed model showed significant improvements on short queries and consistent performance on other queries. Another planned improvement is to retrain the model to cover new brands and new product types recently being added to our product catalog. We plan to first add these new brands and product types to the synthetic data (i.e. I.6 in Fig. \ref{fig:flowchart}), which takes minimal effort. 
If the improvement is not satisfactory, we may incrementally add a small amount of such data to the golden data set, which is still additional minimal effort.

\section{Discussion and Future Work} \label{sec:dis}
We discuss our TripleLearn framework to address three questions below. 
\\

\textbf{1) What challenges can TripleLearn help with?}\ TripleLearn is a novel training framework that works with less satisfactory training data. In most industrial settings, it is often too expensive to prepare a large amount of high quality training data. However, it is more realistic to prepare three sets of training data as required by TripleLearn. As shown in the end-to-end process (Fig. \ref{fig:flowchart}), the synthetic data and large volume noisy data can be generated automatically, while the golden data is a small set which can be annotated manually within a reasonable cost. 

Meanwhile, separate datasets are also easier to maintain in practice because they are independent, and we may only update one set of training data. For example, an eCommerce business adds or removes products regularly. Using TripleLearn, only the synthetic data has to be updated to reflect the latest product catalog.
\\

\textbf{2) Why TripleLearn works?} \ TripleLearn leverages three separate datasets that provide collaborative supervision. Golden data is the ground truth to start the model training and to measure the real model performance; the large noisy data provides rich variations of real-world patterns; and the synthetic data feeds the model all possible entity values which may or may not show up in the golden data or the large noisy data.

TripleLearn may resemble a natural way of learning. An analogy is the process of learning a new language. Usually, we may learn from three types of materials:  1) a textbook with high quality content, exercises, and solutions; 2) a dictionary with all the words and common phrases;  3) a variety of real-world materials such as TV shows and movies. These materials may be related to the three datasets for TripleLearn. The high quality golden data corresponds to the textbook; the synthetic data covers all the brands and product types, which is similar to the dictionary; and the large amount of noisy data plays a role as the variety of real-world materials which may be noisy too. 
With these three types of materials, learning a new language is also an iterative process. Specifically, we learn from each type of the materials in each "iteration", then validate our learning using high quality textbook content and exercises. We keep going through this process iteratively so that we can learn more vocabulary and more complex patterns, which is similar to how TripleLearn works. 
\\

\textbf{3) How to generalize TripleLearn?}\  TripleLearn can be generalized to different models and different problems. In terms of the models, TripleLearn is model-independent and can use any type of supervised models. In terms of the problems, it can be applied to any problem with similar training data issues. For example, NER problem in eCommerce is a perfect fit because of the same data foundation and requirements, such as extracting entities from search queries or product titles and descriptions. Another candidate is Machine Translation \cite{koehn2017six} because it usually has well-defined dictionaries but a large amount of noisy training data.

\hfill

In the future, we plan to work further in three directions. Firstly, more entities, such as color and size, will be included in the model. Secondly, the use cases can be extended to offline applications, such as extracting entities from product descriptions to enrich a knowledge graph. Lastly, we would like to test this framework using public available datasets for easier reproducibility. 

\section{Conclusion}

Our work demonstrates an end-to-end solution to apply state-of-art deep learning model in a domain-specific industrial problem, i.e. named entity recognition on eCommerce search queries.  The core is a novel model training framework TripleLearn which can iteratively learn from three separate sets of training data.  We demonstrate that this iterative process is effective at improving model performance, instead of traditionally training using one set of data (Sec. \ref{sec:exptriple}). The resulting model by TripleLearn has been deployed in production as a real-time web service for the search engine at homedepot.com.  The model A/B test and day-to-day use for more than 9 months show stable model performance, higher user engagement, and increased revenue, which proves the significant value in the domain of eCommerce. Moreover, the TripleLearn framework is practically easy to maintain and refresh the model by allowing incrementally update one or multiple of the three datasets (Sec. \ref{sec:prodmaintain}). 

More importantly, as discussed in Sec. \ref{sec:dis}, our proposed TripleLearn framework can be generalized to different models and problems.  We hope that it will inspire more industrial innovations using data science and machine learning.

\bibliography{references}

\begin{thebibliography}{24}
\providecommand{\natexlab}[1]{#1}
\providecommand{\url}[1]{\texttt{#1}}
\providecommand{\urlprefix}{URL }
\expandafter\ifx\csname urlstyle\endcsname\relax
  \providecommand{\doi}[1]{doi:\discretionary{}{}{}#1}\else
  \providecommand{\doi}{doi:\discretionary{}{}{}\begingroup
  \urlstyle{rm}\Url}\fi

\bibitem[{Collobert and Weston(2008)}]{collobert2008}
Collobert, R.; and Weston, J. 2008.
\newblock A unified architecture for natural language processing: Deep neural
  networks with multitask learning.
\newblock In \emph{Proceedings of the 25th international conference on Machine
  learning}, 160--167.

\bibitem[{Collobert et~al.(2011)Collobert, Weston, Bottou, Karlen, Kavukcuoglu,
  and Kuksa}]{collobert2011cnn}
Collobert, R.; Weston, J.; Bottou, L.; Karlen, M.; Kavukcuoglu, K.; and Kuksa,
  P. 2011.
\newblock Natural language processing (almost) from scratch.
\newblock \emph{Journal of machine learning research} 12(Aug): 2493--2537.

\bibitem[{Cowan et~al.(2015)Cowan, Zethelius, Luk, Baras, Ukarde, and
  Zhang}]{cowan15expedia}
Cowan, B.; Zethelius, S.; Luk, B.; Baras, T.; Ukarde, P.; and Zhang, D. 2015.
\newblock Named Entity Recognition in Travel-related Search Queries.
\newblock In \emph{Proceedings of the Twenty-Ninth AAAI Conference on
  Artificial Intelligence}, AAAI'15, 3935--3941. AAAI Press.
\newblock ISBN 0-262-51129-0.
\newblock \urlprefix\url{http://dl.acm.org/citation.cfm?id=2888116.2888261}.

\bibitem[{Devlin et~al.(2018)Devlin, Chang, Lee, and Toutanova}]{bert2018}
Devlin, J.; Chang, M.-W.; Lee, K.; and Toutanova, K. 2018.
\newblock Bert: Pre-training of deep bidirectional transformers for language
  understanding.
\newblock \emph{arXiv preprint arXiv:1810.04805} .

\bibitem[{Guo et~al.(2009)Guo, Xu, Cheng, and Li}]{guo2009nermicrosoft}
Guo, J.; Xu, G.; Cheng, X.; and Li, H. 2009.
\newblock Named entity recognition in query.
\newblock In \emph{Proceedings of the 32nd international ACM SIGIR conference
  on Research and development in information retrieval}, 267--274. ACM.

\bibitem[{Huang, Xu, and Yu(2015)}]{huang2015lstm}
Huang, Z.; Xu, W.; and Yu, K. 2015.
\newblock Bidirectional LSTM-CRF models for sequence tagging.
\newblock \emph{arXiv preprint arXiv:1508.01991} .

\bibitem[{Koehn and Knowles(2017)}]{koehn2017six}
Koehn, P.; and Knowles, R. 2017.
\newblock Six challenges for neural machine translation.
\newblock \emph{arXiv preprint arXiv:1706.03872} .

\bibitem[{Lample et~al.(2016)Lample, Ballesteros, Subramanian, Kawakami, and
  Dyer}]{lample16neural}
Lample, G.; Ballesteros, M.; Subramanian, S.; Kawakami, K.; and Dyer, C. 2016.
\newblock Neural Architectures for Named Entity Recognition.
\newblock \emph{CoRR} abs/1603.01360.
\newblock \urlprefix\url{http://arxiv.org/abs/1603.01360}.

\bibitem[{Lee(2017)}]{lee2017lstm}
Lee, C. 2017.
\newblock LSTM-CRF models for named entity recognition.
\newblock \emph{IEICE Transactions on Information and Systems} 100(4):
  882--887.

\bibitem[{Li et~al.(2020)Li, Sun, Han, and Li}]{li2020survey}
Li, J.; Sun, A.; Han, J.; and Li, C. 2020.
\newblock A survey on deep learning for named entity recognition.
\newblock \emph{IEEE Transactions on Knowledge and Data Engineering} .

\bibitem[{Liu et~al.(2019)Liu, Meng, Zhang, Xu, Chen, and
  Zhou}]{liu2019bertner}
Liu, Y.; Meng, F.; Zhang, J.; Xu, J.; Chen, Y.; and Zhou, J. 2019.
\newblock Gcdt: A global context enhanced deep transition architecture for
  sequence labeling.
\newblock \emph{arXiv preprint arXiv:1906.02437} .

\bibitem[{Ma and Hovy(2016)}]{ma2016cnn}
Ma, X.; and Hovy, E. 2016.
\newblock End-to-end sequence labeling via bi-directional lstm-cnns-crf.
\newblock \emph{arXiv preprint arXiv:1603.01354} .

\bibitem[{Majumder et~al.(2018)Majumder, Subramanian, Krishnan, Gandhi, and
  More}]{majumder2018walmart}
Majumder, B.~P.; Subramanian, A.; Krishnan, A.; Gandhi, S.; and More, A. 2018.
\newblock Deep Recurrent Neural Networks for Product Attribute Extraction in
  ECommerce.
\newblock \emph{arXiv preprint ArXiv:1803.11284} .

\bibitem[{Mikolov et~al.(2013)Mikolov, Chen, Corrado, and
  Dean}]{mikolov2013efficient}
Mikolov, T.; Chen, K.; Corrado, G.; and Dean, J. 2013.
\newblock Efficient estimation of word representations in vector space.
\newblock \emph{arXiv preprint arXiv:1301.3781} .

\bibitem[{More(2016)}]{more16walmart}
More, A. 2016.
\newblock Attribute Extraction from Product Titles in eCommerce.
\newblock \emph{CoRR} abs/1608.04670.
\newblock \urlprefix\url{http://arxiv.org/abs/1608.04670}.

\bibitem[{Nadeau and Sekine(2007)}]{sekine09nyu}
Nadeau, D.; and Sekine, S. 2007.
\newblock A survey of named entity recognition and classification.
\newblock \emph{Lingvisticæ Investigationes} 30(1): 3--26.
\newblock ISSN 0378-4169.
\newblock \doi{https://doi.org/10.1075/li.30.1.03nad}.
\newblock
  \urlprefix\url{https://www.jbe-platform.com/content/journals/10.1075/li.30.1.03nad}.

\bibitem[{Pennington, Socher, and Manning(2014)}]{glove2014}
Pennington, J.; Socher, R.; and Manning, C. 2014.
\newblock Glove: Global vectors for word representation.
\newblock In \emph{Proceedings of the 2014 conference on empirical methods in
  natural language processing (EMNLP)}, 1532--1543.

\bibitem[{Peters et~al.(2018)Peters, Neumann, Iyyer, Gardner, Clark, Lee, and
  Zettlemoyer}]{elmo2018}
Peters, M.~E.; Neumann, M.; Iyyer, M.; Gardner, M.; Clark, C.; Lee, K.; and
  Zettlemoyer, L. 2018.
\newblock Deep contextualized word representations.
\newblock \emph{arXiv preprint arXiv:1802.05365} .

\bibitem[{Putthividhya and Hu(2011)}]{ner2011ebay}
Putthividhya, D.~P.; and Hu, J. 2011.
\newblock Bootstrapped Named Entity Recognition for Product Attribute
  Extraction.
\newblock In \emph{Proceedings of the Conference on Empirical Methods in
  Natural Language Processing}, EMNLP '11, 1557--1567. Stroudsburg, PA, USA:
  Association for Computational Linguistics.
\newblock ISBN 978-1-937284-11-4.
\newblock \urlprefix\url{http://dl.acm.org/citation.cfm?id=2145432.2145598}.

\bibitem[{Sang and De~Meulder(2003)}]{conll2003}
Sang, E.~F.; and De~Meulder, F. 2003.
\newblock Introduction to the CoNLL-2003 shared task: Language-independent
  named entity recognition.
\newblock \emph{arXiv preprint cs/0306050} .

\bibitem[{Vaswani et~al.(2017)Vaswani, Shazeer, Parmar, Uszkoreit, Jones,
  Gomez, Kaiser, and Polosukhin}]{vaswani2017attention}
Vaswani, A.; Shazeer, N.; Parmar, N.; Uszkoreit, J.; Jones, L.; Gomez, A.~N.;
  Kaiser, {\L}.; and Polosukhin, I. 2017.
\newblock Attention is all you need.
\newblock In \emph{Advances in neural information processing systems},
  5998--6008.

\bibitem[{Wen et~al.(2019)Wen, Vasthimal, Lu, Wang, and Guo}]{wen2019ebay}
Wen, M.; Vasthimal, D.~K.; Lu, A.; Wang, T.; and Guo, A. 2019.
\newblock Building Large-Scale Deep Learning System for Entity Recognition in
  E-Commerce Search.
\newblock In \emph{Proceedings of the 6th IEEE/ACM International Conference on
  Big Data Computing, Applications and Technologies}, 149--154.

\bibitem[{Wu et~al.(2017)Wu, Ahmed, Kumar, and Datta}]{wu2017google}
Wu, C.-y.; Ahmed, A.; Kumar, G.~R.; and Datta, R. 2017.
\newblock Predicting Latent Structured Intents from Shopping Queries.
\newblock \emph{WWW 2017} .

\bibitem[{Yadav and Bethard(2019)}]{yadav2019survey}
Yadav, V.; and Bethard, S. 2019.
\newblock A survey on recent advances in named entity recognition from deep
  learning models.
\newblock \emph{arXiv preprint arXiv:1910.11470} .

\end{thebibliography}

\end{document}